%% file: main.tex
\documentclass[sigconf]{acmart}

\usepackage{epsfig}
\usepackage{graphicx}
\usepackage{amsmath}

\usepackage[linesnumbered,ruled]{algorithm2e}
\usepackage{subfig}

\usepackage{comment}
\usepackage{float}
\usepackage{mathtools}
\usepackage{cuted}
\usepackage{xcolor}
\usepackage{array}
\usepackage{makecell}
\usepackage{arydshln}
\usepackage{multirow}
\usepackage{tabularx}
\usepackage{stfloats}
\usepackage{enumitem}

\copyrightyear{2022}
\acmYear{2022}
\setcopyright{rightsretained}
\acmConference[GECCO '22]{Genetic and Evolutionary Computation
Conference}{July 9--13, 2022}{Boston, MA, USA}
\acmBooktitle{Genetic and Evolutionary Computation Conference (GECCO
'22), July 9--13, 2022, Boston, MA, USA}
\acmDOI{10.1145/3512290.3528694}
\acmISBN{978-1-4503-9237-2/22/07}

\begin{document}
\newcommand\ourmethod{Ours}
\newcommand\revision{\color{red}}

\title{Evolving Transferable Neural Pruning Functions}


\author{Yuchen Liu}
\affiliation{
    \institution{Princeton University}
}
\email{yl16@princeton.edu}

\author{S.Y. Kung}
\affiliation{
    \institution{Princeton University}
}
\email{kung@princeton.edu}

\author{David Wentzlaff}
\affiliation{
    \institution{Princeton University}
}
\email{wentzlaf@princeton.edu}

\renewcommand\shortauthors{Liu et al.}

\input{abstract}

\maketitle

\input{introduction}
\input{related_work}
\input{methodology}
\input{co-evolution}

\input{transfer-pruning}

\input{ablation_study}
\input{conclusion}

\input{acknowledgement}

\bibliographystyle{Sty_Ref/ACM-Reference-Format}
\bibliography{Sty_Ref/refs}

\newpage
\input{supplementary}

\end{document}

%% file: abstract.tex
\begin{abstract}

Structural design of neural networks is crucial for the success of deep learning.
While most prior works in evolutionary learning  aim at directly searching the structure of a network,
few attempts have been made on another promising track, 
channel pruning,
which recently has made major headway in designing efficient deep learning models. 
In fact, prior pruning methods adopt human-made pruning functions to score a channel's importance for channel pruning,
which requires domain knowledge and could be sub-optimal. 
To this end, 
we pioneer the use of genetic programming (GP) to discover strong pruning metrics automatically.
Specifically, we craft a novel design space to express high-quality and transferable pruning functions,
which ensures an end-to-end evolution process 
where no manual modification is needed on the evolved functions for their transferability after evolution.
Unlike prior methods,
our approach can provide both compact pruned networks for efficient inference
and novel closed-form pruning metrics which are mathematically explainable and thus generalizable to different pruning tasks.
While the evolution is conducted on small datasets,
our functions shows promising results when applied to more challenging datasets, 
different from those used in the evolution process. 
For example, on ILSVRC-2012,
an evolved function achieves state-of-the-art pruning results.
\end{abstract}

%% file: introduction.tex
\section{Introduction}
\label{sec:intro}

Convolutional neural networks (CNNs) have demonstrated superior performance on various computer vision tasks~\cite{deng2009imagenet,goodfellow2014generative,ren2015faster,dong2015image}.
However, CNNs require huge storage space, high computational budget, and large memory utilization, 
which could far exceed the resource limit of edge devices like mobile phones and embedded gadgets.
As a result,
many methods have been proposed to reduce their cost, such as 
weight quantization~\cite{chen2015compressing, courbariaux2016binarized,han2015deep}, 
tensor factorization~\cite{jaderberg2014speeding, lebedev2014speeding},  
weight pruning~\cite{han2015learning,zhang2018systematic},
and channel pruning~\cite{ he2019filter,liu2021content}. 
Among them all, channel pruning is the preferred approach to learn dense compact models, which has been receiving increased focus from the research community. 

Channel pruning is usually achieved in three steps: 
(1) score channels' importance with a hand-crafted pruning function; 
(2) remove redundant channels based on the scores;
(3) retrain the network. 
The performance of channel pruning largely depends on the pruning function used in step (1). 
Current scoring metrics are mostly handcrafted to extract crucial statistics from channels' feature maps~\cite{he2017channel,yu2018nisp} or kernel parameters~\cite{li2016pruning,he2019filter} 
in a labelless~\cite{liu2017learning,he2018soft} or label-aware~\cite{zhuang2018discrimination,kung2019methodical} manner.
However, the design space of pruning functions is so large that 
hand-crafted metrics are usually sub-optimal, 
and enumerating all functions with human labor under the space is impossible.
While prior evolutionary learning works aim to automate the design for the structure of the network directly~\cite{suganuma2017genetic,sinha2021evolving},
no attempts, to the best of our knowledge, have been made to evolve the pruning metrics.
These raise the question: can we leverage evolutionary strategies to automatically develop strong pruning functions to advance channel pruning?

\begin{figure*}[t]
\centering
\includegraphics[width=\textwidth]{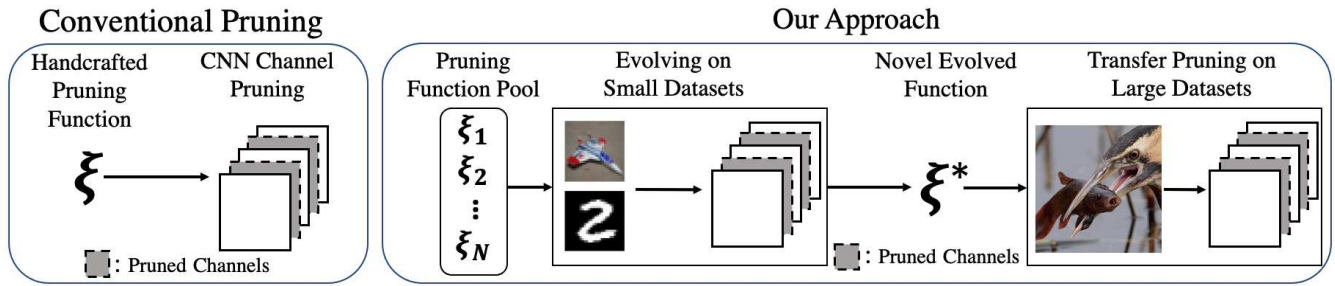}
\vspace{-0.7cm}
\caption{Illustration of our approach. 
Compared to conventional methods which mainly use handcrafted pruning functions,
we aim to learn the pruning functions automatically via an evolution strategy, genetic programming.
The evolved functions are transferable and generalizable, further enhancing the pruning performance.
}~\label{fig:general_idea}
\end{figure*}

To this end, we take the first step to adopt genetic programming (GP) to learn transferable pruning functions, 
as shown in Fig.~\ref{fig:general_idea}. 
In particular, a population of functions is evolved by applying them to pruning tasks of small image classification datasets, 
and the evolved functions can later be transferred to larger and more challenging datasets.
Our closed-form, explainable, learned functions are transferable and generalizable:
(1) They are applicable to pruning tasks of different image datasets and networks, and can also be used for other machine learning tasks, e.g., feature selection;
(2) they demonstrate competitive pruning performance on datasets and networks that are different from those used in the evolution process.
Such transferability and generalizability provides a unique advantage to our method, 
where prior meta-pruning methods like MetaPruning~\cite{liu2019metapruning} and LFPC~\cite{he2020learning} are learned and evolved on the same tasks with no transferability and perform inferior to our approach.

Specifically, we encode pruning functions using expression trees
where we carefully design our search space to allow transferability of the evolved functions.
For example, we propose a uni-tree search space for label-aware pruning metrics, which makes them applicable to different datasets.
Such a design space ensures an end-to-end evolution process,
where the learned functions are transferable to other datasets without any manual modification after evolution.
Moreover, under our encoding space, we are able to  build a group of competitive hand-crafted pruning functions, 
which we name as SOAP (state-of-the-art population),  
and we find the use of SOAP considerably improves the evolution performance.
The populations of the functions are evolved with two different pruning tasks, 
LeNet on MNIST and VGGNet on CIFAR-10.
We observe that evolving on two tasks produces better functions than only evolving on one of them,
and more surprisingly, our scheme can even produce more effective pruning functions than direct evolution on a large dataset, 
e.g., ILSVRC-2012, under the same computational budget.
We analyze the merits of an evolved function both mathematically and visually
and transfer it to three larger datasets, CIFAR-100, SVHN, and ILSVRC-2012, 
where it exceeds the state-of-the-art pruning results on all of them.

Our main contributions are summarized as follows:
\begin{itemize}
    \item  We propose a novel paradigm where we leverage genetic programming to learn transferable channel pruning functions which advance pruning efficacy.
    \item We develop a novel design space to allow an end-to-end co-evolution process for searching transferable pruning functions. 
    Such a space also enables us to express SOAP, which helps improve the effectiveness of the evolution.
    \item We provide an analysis on our closed-form evolved functions, 
    which could further streamline the design of pruning metrics.
    The evolved functions also show generalizability to other machine learning tasks, e.g., feature selection.
    \item When transferred to datasets unseen by the evolution,
    our evolved functions achieve state-of-the-art pruning results.
    For example, with 26.9\% and 53.4\% FLOPs\footnote{Number of floating points operations for an image inference.} reduction from MobileNet-V2, 
    we achieve top-1 accuracies of 71.90\% and 69.16\% on ILSVRC-2012, outperforming the state of the art.
\end{itemize}



%% file: related_work.tex
\section{Related Work} \label{sec:related_work}

{\bf Hand-Crafted Channel Pruning.} 
Channel pruning~\cite{yu2018nisp,kung2019methodical,zhuang2018discrimination,he2019filter} is generally realized by using a handcrafted pruning function to score channels' saliency and remove redundant ones. 
Based on the scoring procedure, 
it can be categorized into 
labelless pruning and label-aware pruning.

Labelless channel pruning typically adopts the norm-based property of 
the channel's feature maps or associated filters as pruning criterion~\cite{li2016pruning, liu2017learning, he2017channel, luo2017thinet, louizos2017learning, he2018soft,ye2018rethinking, he2019filter,li2019exploiting}.
For example, 
Liu et al.~\cite{liu2017learning} and Ye et al.~\cite{ye2018rethinking} use the absolute value of scaling factors in the batch-norm,
while $\ell$1-norm and $\ell$2-norm of 
channels' associated filters are computed in~\cite{li2016pruning,he2018soft,li2019exploiting} as channels' importance. 
On the other hand, researchers have designed metrics to 
evaluate class discrepancy of channels' feature maps for label-aware pruning~\cite{zhuang2018discrimination,kung2019methodical,liu2020rethinking}.
Zhuang et al.~\cite{zhuang2018discrimination} insert discriminant losses in the network and 
remove channels that are the least correlated to the losses after iterative optimization.
Kung et al.~\cite{kung2019methodical} and Liu et al.~\cite{liu2020rethinking, liu2021class} adopt closed-form discriminant functions to accelerate the scoring process. 

While these works use handcrafted scoring metrics,
we learn transferable and generalizable pruning functions automatically.  

\noindent\textbf{Meta-Learning.} Our work falls into the category of meta-learning, 
where prior works have attempted to optimize machine learning components, 
including 
hyper-parameters~\cite{bergstra2013making, snoek2015scalable,feurer2015efficient}, 
optimizers~\cite{chen2017learning, wichrowska2017learned,bello2017neural}, 
and neural network structures~\cite{zoph2016neural,zoph2018learning,tan2019efficientnet,liu2018progressive,xie2017genetic,real2017large,liu2017hierarchical,real2019regularized}.

Prior works on neural architecture search (NAS) have leveraged reinforcement learning (RL) to discover high-performing network structures~\cite{zoph2016neural,baker2016designing,zoph2018learning,cai2018proxylessnas,tan2019mnasnet,tan2019efficientnet}.
Recently, NAS has also been adopted to find efficient network structures~\cite{tan2019efficientnet,tan2019mnasnet}.
Another line of works adopts evolution strategies (ES) to explore the space of network structures~\cite{fernando2016convolution,
suganuma2017genetic,
xie2017genetic,real2017large,
liu2017hierarchical,
real2019regularized,
dai2019chamnet,
miikkulainen2019evolving,
stanley2019designing,
o2020neural,
sinha2021evolving, templier2021geometric},
which demonstrates competitive performance to RL methods.
This notion is pioneered by neuro-evolution~\cite{stanley2002evolving,floreano2008neuroevolution, stanley2009hypercube}
which evolves the topology of small neural networks.
In the era of deep learning, Suganuma et al.~\cite{suganuma2017genetic} leverage Cartesian genetic programming to find competitive network structures.
Real et al.~\cite{real2019regularized} evolve networks that improve over the ones found by RL-based NAS~\cite{zoph2018learning}.
Dai et al.~\cite{dai2019chamnet} apply ES to design efficient and deployable networks for mobile platforms.
Templier et al.~\cite{templier2021geometric} propose a geometric encoding scheme for more efficient parameter search.

Compared to prior works, 
we employ evolutionary learning from a new angle for efficient network design, 
where we learn transferable pruning functions that produce state-of-the-art pruning results.
Our work is orthogonal to prior works, for example, our evolved functions can be potentially applied on evolutionary NAS-learned networks to further enhance their efficiency.

\noindent\textbf{Meta-Pruning.} 
Prior works~\cite{huang2018learning,he2018amc,liu2019metapruning,chin2020towards,he2020learning} have also adopted a similar notion of learning to prune a CNN. 
We note that an evolution strategy is used in LeGR~\cite{chin2020towards} and MetaPruning~\cite{liu2019metapruning} 
to search for a pair of pruning parameters and network encoding vectors, respectively.
However,
our approach is drastically different from them in terms of search space and search candidates,
where we search for effective combinations of operands and operators  to build transferable and powerful pruning functions.
He et al. propose LFPC~\cite{he2020learning} to learn network pruning criteria (functions) across layers by training a differentiable criteria sampler.
However, rather than learning new pruning functions, 
their goal is to search within a pool of existing pruning criteria and find candidates that are good for a certain layer's pruning.
On the contrary, 
our evolution recombines the operands and operators 
and produces novel pruning metrics, 
which are generally good for all layers.

We also notice that MetaPruning~\cite{liu2019metapruning}, LFPC~\cite{he2020learning}, and other methods~\cite{huang2018learning,he2018amc,chin2020towards} are all learned on one task (dataset and network) and applied only on the same task with no transferability.
In contrast, we only need one evolution learning process, which outputs evolved functions 
that are transferable across multiple tasks and demonstrate competitive performance on all of them.

%% file: methodology.tex
\section{Methodology} \label{sec:methodology}

\begin{figure*}[t]
    \centering
    \includegraphics[width=\textwidth]{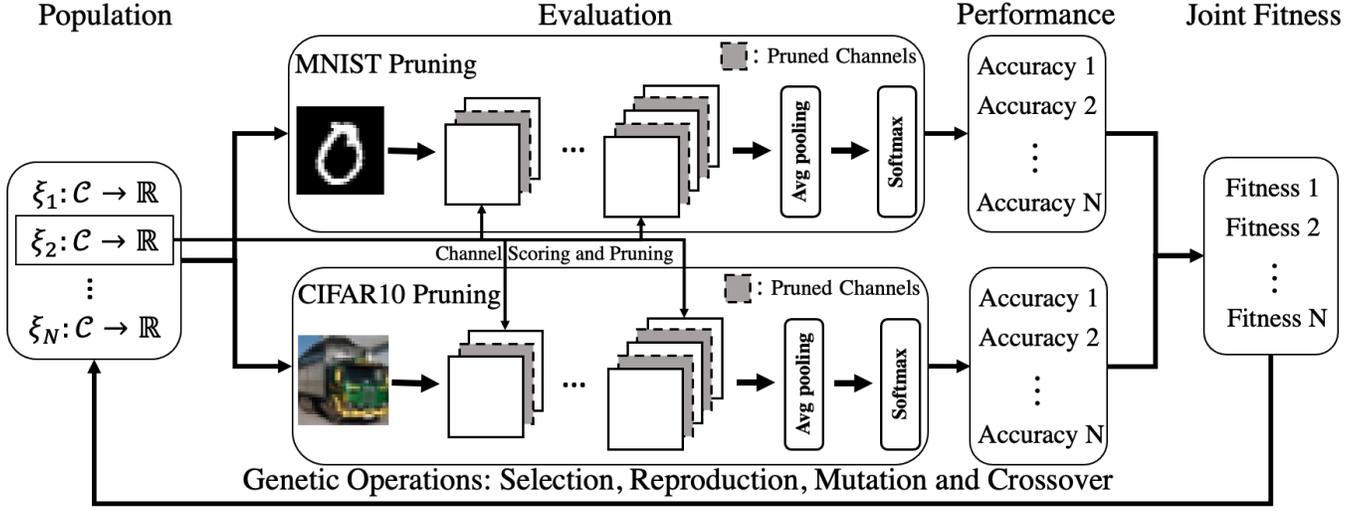}
    \vspace{-0.7cm}
    \caption{ Illustration of our approach to evolve channel pruning functions. 
    A population of functions is applied to conduct pruning tasks on two datasets, MNIST and CIFAR-10.
    Each function receives a fitness value by combining its pruned networks' accuracies. 
    The population will then go through a natural selection process to improve the functions' effectiveness.
    }
    \label{fig:overall_flowchart}
\end{figure*}

In Fig.~\ref{fig:overall_flowchart},
we present our evolution framework,
which leverages
genetic programming~\cite{koza1992genetic} to 
learn high-quality channel pruning functions.
We first describe the design space to encode channel scoring functions.
Next, we discuss the pruning tasks to evaluate the functions' effectiveness.
Lastly, genetic operators are defined to traverse the function space for competitive solutions.

\subsection{Function Design Space }

\input{Table/function_design_space}

{\bf Expression Tree.} In channel pruning, a pruning function $\xi : \mathcal{C} \longmapsto     \mathbb{R}$ scores the channels to determine their importance/redundancy,
where $\mathcal{C}$ denotes feature maps, filters, and their statistics associated with the channels.
This scoring process can be viewed as a series of operations with operators (addition, matrix multiplication, etc.) and operands (feature maps, filters, etc.).
We thus adopt an expression tree encoding scheme to represent a pruning function
where inner nodes are operators, and leaves are operands. 

As shown in Tab.~\ref{tab:operand_space} and~\ref{tab:operator_space}, our function design space includes two types of operands (6 operands in total) 
and four types of operators (23 operators in total), 
via which a vast number of pruning functions can be expressed.
The statistics operators can compute the statistics of an operand in two dimensions, 
namely, global dimension (subscript with `g') and sample dimension (subscript with `s'). 
The global
dimension operators flatten operands into a 1D sequence and extract corresponding statistics, 
while the sample dimension operators compute statistics on the axis of samples.
For example, $\mathrm{sum}_g(\mathcal{W})$ returns the summation of all entries of a kernel tensor, 
while $\mathrm{mean}_s(\mathcal{F})$ returns $\bar{f} \in \mathbb{R}^{H \times W}$, which is the sample average of all feature maps.
We also include specialized operators 
which allow us to build complicated but competitive metrics like maximum mean discrepancy (MMD)~\cite{gretton2012kernel} and filter's geometric median~\cite{he2019filter}.

\begin{figure*}[t]
\centering
\includegraphics[width=0.7\textwidth]{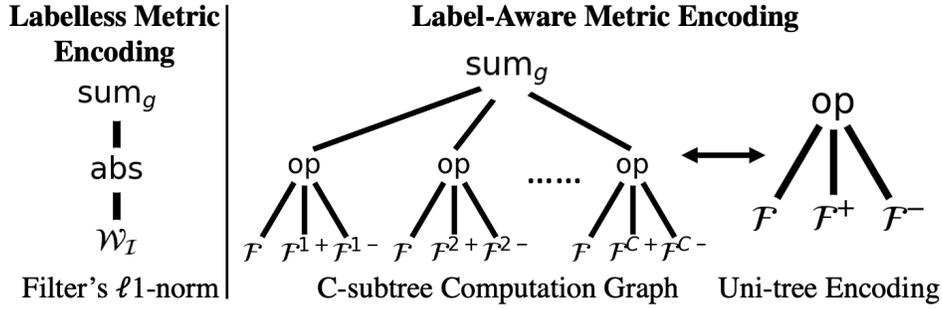}
\vspace{-0.25cm}
\caption{Illustration of the pruning function encoding.
\textbf{Left}: For labelless scoring metrics like filter's $\ell$1-norm, 
we adopt a direct tree encoding scheme.
\textbf{Right}: For label-aware scoring metrics,  
we encode the $C$-subtree computation graph by a uni-tree ($C$: number of class labels). 
The uni-tree encodes the common operations (\textit{op}) on each label partition ($\mathcal{F}^+, \mathcal{F}^-$) and all feature maps ($\mathcal{F}$).
This scheme allows transferable function evolution.
}\label{fig:func_encoding}
\end{figure*}

\noindent{\bf Function Encoding.} 
The channel scoring functions can be categorized into two types: 
labelless metrics and label-aware metrics.
For labelless functions like filter's $\ell 1$-norm, 
we adopt a direct encoding scheme as 
$\mathrm{sum}_g(\mathrm{abs}(\mathcal{W}_I))$
with the expression tree shown in Fig.~\ref{fig:func_encoding}. 

For label-aware metrics such as the one in~\cite{kung2019methodical} and MMD~\cite{gretton2012kernel},
which measure class discrepancy of the feature maps,
we observe a common computation graph among them, as shown in Fig.~\ref{fig:func_encoding}:
(1)~partition the feature maps in a labelwise manner;
(2) apply the same operations on each label partition and all feature maps; 
(3)  average/sum the scores of all partitions to obtain a single scalar.
These metrics can be naively encoded as $C$-branch trees ($C$: number of class labels in the dataset).
However, directly using the naive encoding scheme will result in data-dependent non-transferable metrics because: 
(1) $C$ varies from dataset to dataset (e.g., metrics for CIFAR-10 is not transferable to CIFAR-100); 
(2) mutating the subtrees differently could make the metric overfit to a specific label numbering scheme 
(e.g., for a metric with different subtrees on class-1 and class-2, 
renumbering the labels would mean the metric would compute something different, which is undesirable).

To combat the above issues, 
we express a label-aware function by a uni-tree which encodes the 
common operations that are applied to each label partition, as explained in Fig.~\ref{fig:func_encoding}.
Instead of directly encoding the operands from a specific label partition, 
like $\mathcal{F}^{1+}$ (feature maps with labels equal to 1) and $\mathcal{F}^{1-}$ (feature maps with labels not equal to 1), 
we use a symbolic representation of $\mathcal{F^+}$ and $\mathcal{F^-}$ to generically encode the partition concept.
In the actual scoring process, 
the uni-tree is compiled back to a $C$-branch computation graph, 
with $\mathcal{F^+}$ and $\mathcal{F^-}$ converted to the specific map partitions.
Such uni-tree encoding allows us to evolve label-aware metrics independent of $C$ and label numbering schemes, 
which ensures their transferability to datasets unseen by the evolution process. 

\noindent\textbf{SOAP.} Using the above described function encoding, 
we can implement a broad range of competitive pruning functions:
filter's $\ell1$-norm~\cite{li2016pruning},
filter's $\ell2$-norm~\cite{he2018soft},
batch norm's scaling factor~\cite{liu2017learning}, 
filter's geometric median~\cite{he2019filter},
Discriminant Information~\cite{kung2019methodical},
Maximum Mean Discrepancy~\cite{gretton2012kernel},
Absolute SNR~\cite{golub1999molecular},
Student's T-Test~\cite{lehmann2006testing},
Fisher Discriminant Ratio~\cite{pavlidis2001gene},
and Symmetric Divergence~\cite{mak2006solution}.
For the last four metrics, we adopt the scheme in~\cite{liu2020rethinking} for channel scoring. 
We name this group of functions the state-of-the-art population (SOAP), 
which helps our evolution in many aspects. 
For instance, in Sec.~\ref{sec:ablation_study}, we find that initializing the population with SOAP evolves better pruning functions than random initialization. 
Detailed implementation of SOAP is included in Supplementary.

\subsection{Function Effectiveness Evaluation}

The encoded functions are applied to empirical pruning tasks to evaluate their effectiveness.
To avoid overfitting on certain data patterns and increase the generality of the evolved functions, 
we evolve the population of functions on two different pruning tasks, LeNet-5~\cite{lecun1998gradient} on MNIST~\cite{lecun1998gradient} and VGG-16~\cite{simonyan2014very} on CIFAR-10~\cite{krizhevsky2009learning}.
In both pruning tasks, we adopt a one-shot pruning scheme and report the retrained accuracies on validation sets.
For each pruning task, we keep the pruning settings (layers' pruning ratios, target pruning layers, etc.)
and the retraining hyper-parameters (learning rate, optimizer, weight decay factor, etc.) the same for all evaluations
throughout the evolution process.
This guarantees a fair effectiveness comparison over different functions in all generations 
and ensures we are evolving better functions rather than better hyper-parameters. 
In this way, we can meta-learn powerful functions that perform well on both MNIST and CIFAR-10 and are generalizable to other datasets.
Not surprisingly, evolving with both tasks produce stronger pruning functions than evolving on only one of them, shown in Sec.~\ref{sec:fitness_combination}.
Moreover, in Sec.~\ref{sec:ablation_study}, we find our strategy enjoys better cost-effectiveness compared to direct evolution on a large dataset, e.g., ILSVRC-2012.

\subsection{Function Fitness}\label{sec:fitness_combination}

After evaluation, each encoded function receives two accuracies, $\mathrm{Acc_{MNIST}}$ and $\mathrm{Acc_{CIFAR}}$, from the pruning tasks.
We investigate two accuracy combination schemes,
weighted arithmetic mean (Eqn.~\ref{eqn:arthimetic}) and weighted geometric mean (Eqn.~\ref{eqn:geometric}), 
to obtain the joint fitness of a function.
A free parameter $\alpha \in [0, 1]$ is introduced to control the weights of different tasks.
  \begin{equation}
    \label{eqn:arthimetic}
      \mathrm{Fitness} = \alpha \times \mathrm{Acc_{MNIST}} + (1 - \alpha) \times \mathrm{Acc_{CIFAR}}
  \end{equation} 
 \vspace{-0.6cm}
  \begin{equation}
    \label{eqn:geometric}
 \mathrm{Fitness} = (\mathrm{Acc_{MNIST}})^{\alpha} \times (\mathrm{Acc_{CIFAR}})^{1 - \alpha}  
 \end{equation}

\begin{figure*}[t]
\begin{minipage}{0.4\textwidth}
\centering
\includegraphics[width=0.95\textwidth]{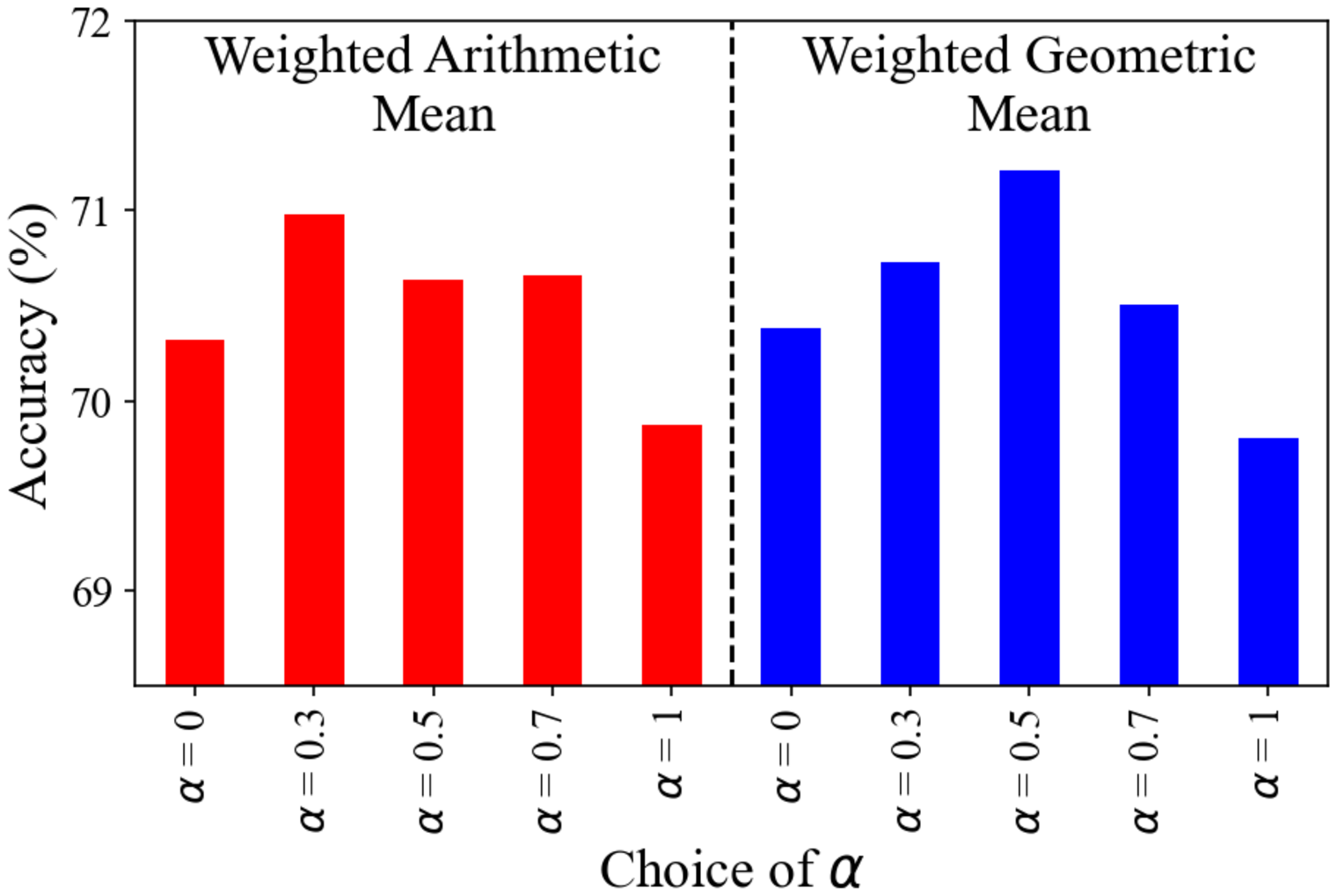}
\vspace{-0.3cm}
\caption{Preliminary evolution tests on the choice of fitness combination scheme. 
The best evolved function from each scheme is applied to conduct a pruning test on CIFAR-100 with ResNet-38, and their accuracies are plotted.}
\label{fig:combination_scheme}
\end{minipage}
\hfill
\begin{minipage}{0.57\textwidth}
\centering
\includegraphics[width=0.8\textwidth]{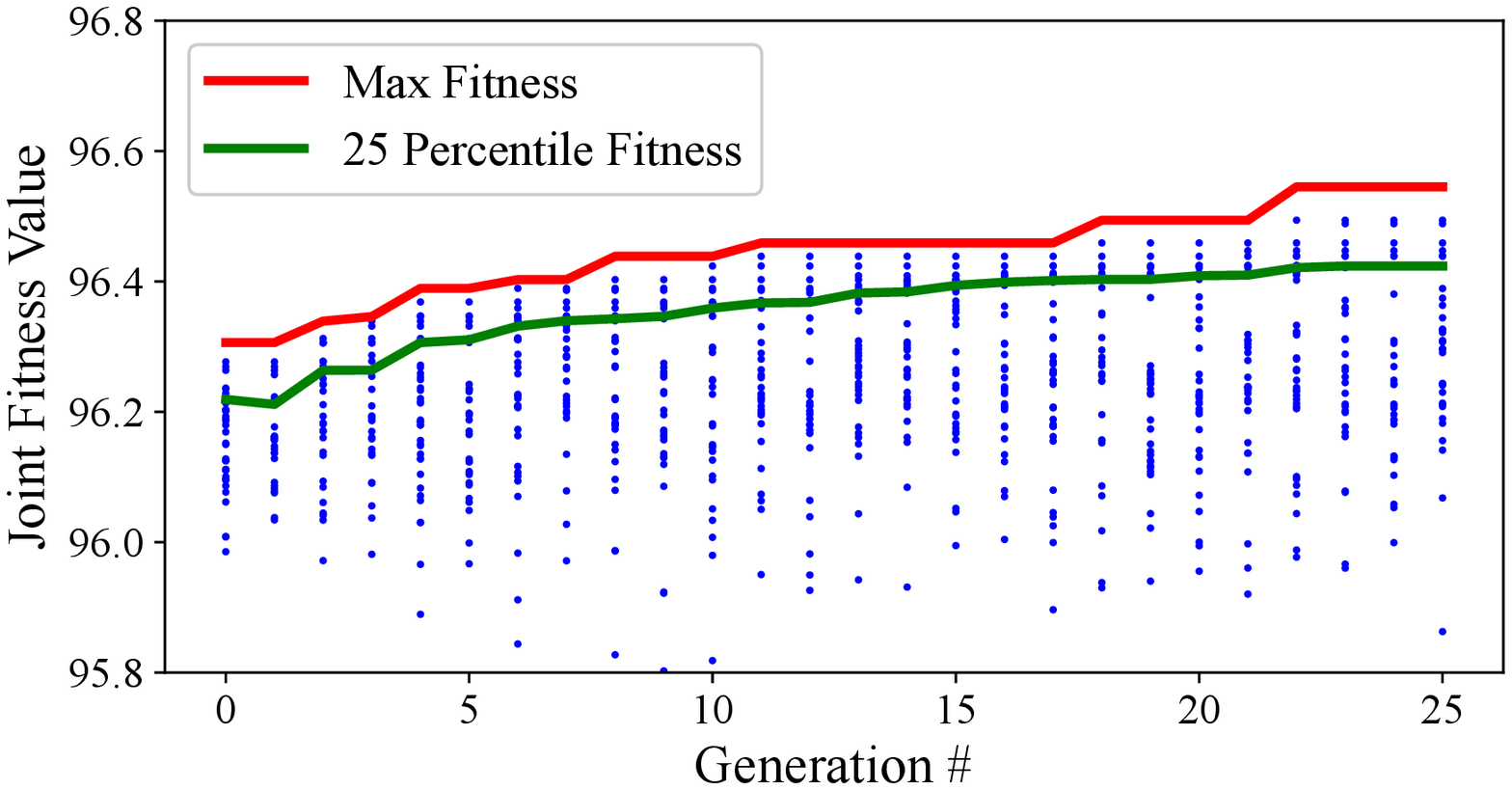}
\vspace{-0.3cm}
\caption{Progress of the evolution experiment. Each dot represents an individual function evaluation. 
The red curve shows functions with the best fitness over generations, 
while the green curve shows the individuals at the 25 percentile fitness.
The effectiveness of the best function and the population's overall quality are both monotonically increasing.
}
\label{fig:evo_results}
\end{minipage}
\end{figure*}

\noindent\textbf{Ablation Study.} 
To decide the fitness combination scheme for the main experiments,
we conduct 10 small preliminary evolution tests
using a grid  of $\alpha \in \{0, 0.3, 0.5, 0.7, 1\}$ with both combination schemes.
Note that when $\alpha \in \{0, 1\}$, the process degenerates to single dataset evolution.
We empirically evaluate the generalizability of the best evolved functions from 
each test by applying them to prune a ResNet-38 on CIFAR-100. 
Note CIFAR-100 is not used in the evolution process, and thus the performance on it speaks well for evolved functions' generalizability.
In Fig.~\ref{fig:combination_scheme}, 
we find that solely evolving on MNIST ($\alpha = 1$) would be the least effective 
option for CIFAR-100 transfer pruning.
In addition, we find that functions evolved on two datasets ($\alpha \in [0.3, 0.5, 0.7]$) 
generally perform better than the ones that just evolve on a single dataset ($\alpha \in [0, 1]$).
We observe that setting
$\alpha = 0.5$ with weighted geometric mean leads to the best result,
which we adopt in the main experiments.

\subsection{Genetic Operations}

\textbf{Selection.} After evaluation, the population will undergo a selection process, 
where we adopt tournament selection~\cite{goldberg1991comparative} to choose a subset of competitive functions. 

\noindent\textbf{Diversity Maintenance.} This subset of functions is then used to reproduce individuals for the next generation.
However,
we observe shrinkage of the population's genetic diversity when all children are reproduced from parents, 
as the selected parents only represent a small pool of genomes.
Such diversity shrinkage would result in premature convergence of the evolution process.
To combat this issue, 
we reserve a slot in the next generation and produce individuals in the slots by randomly cloning functions from SOAP or building random trees.
We find this adjustment empirically useful to help the evolution proceed longer.

\noindent\textbf{Mutation and Crossover.} 
We conduct mutation and crossover on the reproduced population to traverse the function design space for new expressions. 
We adopt the conventional scheme of random tree mutation and one point crossover~\cite{banzhaf1998genetic}.
After mutation and crossover, the population will go through the next evolution iteration.

\noindent\textbf{Function Validity.} 
The function expressions generated from mutation and crossover can be invalid (non-invertible matrix, dimension inconsistency, etc.) due to the random selections of operators, operands, and nodes in the expression trees. 
To combat this issue and enlarge our valid function space, 
some operators are deliberately modified from their standard definition. 
For instance, whenever we need to invert a positive semi-definite scatter matrix $S$, we automatically add a ridge factor $\rho I$, and invert the matrix $S + \rho I$. 
For dimension inconsistency in elementwise operations, we have two options to pad the operand with a smaller dimension: 
(1) with 0 for $+$ and $-$, and 1 for $\times$, and $\div$,
(2) with its own value if it is a scalar.
Moreover, 
we conduct a validity test on the mutated/crossovered functions every time after the mutation/crossover process.
The invalid expressions are discarded, and the mutation/crossover operations are repeated until we recover the population size with all valid functions.
These methods ensure we generate valid function expressions under our vast design space during the evolution process.


%% file: Table/function_design_space.tex
\begin{table*}[t]
    \centering
    
    \begin{tabular}{ | m{10.5em} | m{12.5cm}|}
    
    \hline
    {\bf Filter-based operands} & whole layer's filter $\mathcal{W} \in \mathbb{R}^{c_{out} \times c_{in} \times h \times w}$, channel's incoming filter $\mathcal{W}_I \in \mathbb{R}^{c_{in} \times h \times w}$, channel's batch-normed parameter $\mathcal{B}  \in \mathbb{R}^4$  \\ 
    \hline
    {\bf Map-based operands} &  Feature maps collection $\mathcal{F}= \{ (f_i, y_i) | f_i\in\mathbb{R}^{H \times W}, y_i\in[1:C], i\in[1:N] \}$, two partitions of feature maps collections $\mathcal{F^+} = \{f_i | y_i=k,  k\in[1:C]\}$ and $\mathcal{F^-}=\{f_i|y_i\neq k,  k\in[1:C]\}$
    \\ 
    \hline
    
    \end{tabular}
    \caption{Operand Space}
    \label{tab:operand_space}

    \begin{tabular}{ | m{10.5em} | m{12.5cm}|}
    
    \hline
    {\bf Elementwise operators} & addition, subtraction, multiplication, division, absolute value, square, square root, adding ridge factor \\ 
    \hline
    {\bf Matrix operators} & matrix trace, matrix multiplication, matrix inversion, inner product, outer product, matrix/vector transpose  \\ 
    \hline
    {\bf Statistics operators} & summation, product, mean, standard deviation, variance, counting measure\\ 
    \hline
     {\bf Specialized operators} &  rbf kernel matrix getter, geometric median getter, tensor slicer\\ 
    \hline
    
    \end{tabular}
    \caption{Operator Space}
    \label{tab:operator_space}
\end{table*}

%% file: co-evolution.tex
\section{Evolution on MNIST and CIFAR-10}\label{sec:co-evolve}

\input{Table/SVHN_CIFAR_Transfer_Pruning_Table}

\begin{table*}[t]
\centering
\begin{minipage}{\textwidth}
\begin{equation}
\label{eqn:best-func}
\xi^*(\mathcal{C}) = \frac{\mathrm{var}_g(\mathcal{F}^-)}{\mathrm{var}_g(\mathcal{F}^+)} 
+ \frac{\mathrm{var}_g(\mathcal{F}^+)}{\mathrm{var}_g(\mathcal{F}^-)} 
+ \frac{
||\mathrm{std}_g(\bar{f}) \times \mathrm{var}_g(\mathcal{F}^-) \times 
\bar{f} + (\mathrm{var}_g(\mathcal{F}^+) - \mathrm{mean}_g(\mathcal{F}^-))\mathbf{1}
||^2_2}
{\mathrm{var}_g(\mathcal{F}^+) + \mathrm{var}_g(\mathcal{F}^-) }
\end{equation}
\end{minipage}
\end{table*}

{\bf Experiment Settings.}
We conduct the experiment with a population size of 40 individuals over 25 generations. 
The population is initialized with
20 individuals chosen by randomly cloning functions from SOAP 
and 20 random expression trees. 
The size of the selection tournament is 4 and 
we select 10 functions in each generation.
24 individuals are reproduced from the selected functions, while 6 individuals are from SOAP or randomly built.
The mutation and crossover probability are both set to be 0.75.
We prune 92.4\% of FLOPs from a LeNet-5 (baseline acc: 99.26\%) and 63.0\% of FLOPs from a VGG-16 (baseline acc: 93.7\%), respectively. 
Such aggressive pruning schemes help us better identify functions' effectiveness.
We use the weighted geometric mean in Eqn.~\ref{eqn:geometric} to combine two validation accuracies with $\alpha = 0.5$.
Our codes are implemented with DEAP~\cite{DEAP_JMLR2012} and TensorFlow~\cite{abadi2016tensorflow} for the genetic operations and the neural network pruning.
The experiments are carried out on
a cluster with SLURM job scheduler~\cite{yoo2003slurm} for workload parallelization.

\noindent\textbf{Experiment Result.} Our evolution progress is shown in Fig.~\ref{fig:evo_results},
where the red curve denotes the functions with the maximum fitness 
while the green curve plots the ones with the top 25 percentile fitness.
Both curves increase monotonically over generations, 
indicating that the quality of both the best function and the entire population improves over time. 
This demonstrates the effectiveness of our scheme.
Specifically, the best pruned LeNet-5/VGG-16 in the first generation 
have accuracies of 99.15\%/93.55\% while the best accuracies in the last generation are 99.25\%/94.0\%.
As the first generation is initialized with  SOAP functions, 
such results suggest that the algorithm derives metrics that outperform handcrafted functions in SOAP.
The whole evolution takes 98 GPU-days on P100, which is a reasonable amount of computation for modern evolution learning. 
While this is a pioneering work\footnote{Compared to initial works on NAS, which take 2000 GPU-days~\cite{zoph2018learning} and 3000 GPU-days~\cite{real2017large}, 
we are 20/30x faster.}, we envision that future work could further reduce the evolution computation.

\noindent\textbf{Evolved Function.}
We present the winning function in Eqn.~\ref{eqn:best-func}, 
where $\bar{f} = \mathrm{mean}_s(\mathcal{F})$ denotes sample average of the feature maps and $\mathbf{1}$ is a vector with all entries set to be 1.
The first two terms of the function award a high score to channels with class-diverged feature maps whose $\mathrm{var}_g(\mathcal{F}^+)$ or $\mathrm{var}_g(\mathcal{F}^-)$ is significantly smaller than the other.
Channels with these feature maps contain rich class information as it generates distinguishable responses to different classes.
The third term's 
denominator computes the sum of the feature maps variances while
its numerator 
draws statistics from the average feature maps and the distance between $\mathcal{F}^+$ and $\mathcal{F}^-$, which resembles the concept of signal-to-noise ratio. Two points worth mentioning for this function:
(1) it identifies important statistical concepts from human-designed metrics, where it learns from Symmetric Divergence~\cite{mak2006solution} to measure the divergence of class feature maps. 
(2) it contains unique math concepts that are empirically good for channel importance measurement, 
which is shown in the novel statistics combination of the feature maps in the third term's numerator.
Our visual result in Sec.~\ref{sec:ablation_study} shows $\xi^*$ can be further applied to feature selection, 
which represents another machine learning task.

%% file: Table/SVHN_CIFAR_Transfer_Pruning_Table.tex
\begin{table*}[t]
    \fontsize{8.5}{10}\selectfont
    \centering

    \begin{tabular}{|c|c|c|c|c|c|c|c|c|}
    \hline
    Network & Method & Test Acc (\%) & Acc $\downarrow$ (\%) & FLOPs & Pruned (\%) & Parameters & Pruned (\%) \\
    \hline\hline

    
    \multirow{3}{3em}{\thead{ResNet \\ 164} } & SLIM~\cite{liu2017learning} & 98.22 $\rightarrow$ 98.15 & 0.07 & 172M & 31.1 & 1.46M & 14.5\\

    & {\bf \ourmethod} & {\bf 98.22 $\rightarrow$ 98.26} & {\bf -0.04} & {\bf 92M} & {\bf 63.2} & {\bf 0.64M} & {\bf 63.0}\\
    
    \hline
    \end{tabular}
    \caption{SVHN Transfer Pruning Results}\label{tab:SVHN}

    \begin{tabular}{|c|c|c|c|c|c|c|c|c|}
    \hline
    Network & Method & Test Acc (\%) & Acc $\downarrow$ (\%) & FLOPs & Pruned (\%) & Parameters & Pruned (\%) \\
    \hline\hline
    
    
    \multirow{3}{3em}{VGG19} & SLIM~\cite{liu2017learning} & 73.26 $\rightarrow$ 73.48 & -0.22 & 256M & 37.1 & 5.0M & 75.1 \\
    
    & G-SD~\cite{liu2020rethinking} & 73.40 $\rightarrow$ 73.67 & -0.27 & 161M & 59.5 & 3.2M & 84.0 \\ 
    
    & {\bf \ourmethod} & {\bf 73.40 $\rightarrow$ 74.02} & {\bf -0.62} & {\bf 155M} & {\bf 61.0} & {\bf 2.9M} & {\bf 85.5} \\

    \hline\hline

    
    \multirow{5}{3em}{\thead{ResNet \\ 56}} 
    & SFP~\cite{he2018soft} & 71.33 $\rightarrow$ 68.37 & 2.96 & 76M & 39.3 & - & - \\ 
    
    & FPGM~\cite{he2019filter} & 71.40 $\rightarrow$ 68.79 & 2.61 & 59M & 52.6 & - & - \\ 
    
    & LFPC~\cite{he2020learning} & 71.33 $\rightarrow$ 70.83 & 0.58 & 61M & 51.6 & - & - \\ 
    
    & LeGR~\cite{chin2020towards} & 72.41 $\rightarrow$ 71.04 & 1.37 & 61M & 51.4 & - & - \\
    
    & {\bf \ourmethod} & {\bf 72.05 $\rightarrow$ 71.70} & {\bf 0.35} & {\bf 55M} & {\bf 56.2} & {\bf 0.38M} & {\bf 54.9} \\
    
    \hline\hline

    
    \multirow{5}{3em}{\thead{ResNet \\ 110}} & LCCL~\cite{dong2017more} & 72.79 $\rightarrow$ 70.78 & 2.01 & 173M & 31.3 & 1.75M & 0.0 \\
    
    & SFP~\cite{he2018soft} & 74.14 $\rightarrow$ 71.28 & 2.86 & 121M & 52.3 & - & - \\ 
    
    & FPGM~\cite{he2019filter} & 74.14 $\rightarrow$ 72.55 & 1.59 & 121M & 52.3 & - & - \\ 
    
    & TAS~\cite{dong2019network} & 75.06 $\rightarrow$ 73.16 & 1.90 & 120M & 52.6 & - & - \\ 
    
    & {\bf \ourmethod} & {\bf 74.40 $\rightarrow$ 73.85} & {\bf 0.55} & {\bf 111M} & {\bf 56.2} & {\bf 0.77M} & {\bf 55.8} \\
    
    \hline\hline

    
    \multirow{4}{3em}{\thead{ResNet \\ 164}} 
    & LCCL~\cite{dong2017more} & 75.67 $\rightarrow$ 75.26 & 0.41 & 195M & 21.3 & 1.73M & 0.0\\
    
    & SLIM~\cite{liu2017learning} & 76.63 $\rightarrow$ 76.09 & 0.54 & 124M & 50.6 & 1.21M & 29.7\\
    
    & DI~\cite{kung2019methodical} & 76.63 $\rightarrow$ 76.11 & 0.52 & 105M & 58.0 & 0.95M & 45.1\\


    & {\bf \ourmethod} & {\bf 77.15 $\rightarrow$ 77.77} & {\bf -0.62} & {\bf 92M} & {\bf 63.2} & {\bf 0.66M} & {\bf 61.8} \\
    
    \hline
    \end{tabular}
    \caption{CIFAR-100 Transfer Pruning Results}\label{tab:CIFAR-100}
\end{table*}

%% file: transfer-pruning.tex
\section{Transfer Pruning}

\noindent\textbf{Benchmarks.} To show the generalizability of our evolved pruning function,
we apply $\xi^*$ in Eqn.~\ref{eqn:best-func} to more challenging datasets that are not used in the evolution process:
CIFAR-100~\cite{krizhevsky2009learning}, 
SVHN~\cite{netzer2011reading}, 
and ILSVRC-2012~\cite{deng2009imagenet}.
We compare our method with 
metrics from SOAP, e.g., 
L1~\cite{li2016pruning}, FPGM~\cite{he2019filter}, 
G-SD~\cite{liu2020rethinking},
and DI~\cite{kung2019methodical}, 
where $\xi^*$ outperforms all these handcrafted metrics.
We also include other ``learn to prune" methods like Meta~\cite{liu2019metapruning} and
LFPC~\cite{he2020learning} and other state-of-the-art methods like 
DSA~\cite{ning2020dsa} and CC~\cite{li2021towards} for comparison.
The results are summarized in Tab.~\ref{tab:SVHN},~\ref{tab:CIFAR-100}, and~\ref{tab:ILSVRC-2012},
where the accuracies are shown as ``baseline acc. $\rightarrow$ pruned acc." and the numbers for all other methods are copied from their papers.
On ILSVRC-2012,
we report our pruned models at different FLOPs reduction levels and add a suffix specifying their FLOPs pruning ratios (e.g., \ourmethod\ 60\%-pruned).
This is because different prior arts report their compressed models at different rates, 
and we want to make a fair comparison to all of them.
We find that our evolved function achieves state-of-the-art results on all datasets.

\noindent\textbf{Settings.} We adopt a one-shot pruning scheme with a uniform pruning ratio across layers for our transfer pruning 
and use the SGD optimizer with Nesterov Momentum~\cite{nesterov1983method} for retraining.
The weight decay factor and the momentum are set to be 0.0001 and 0.9, respectively.
On SVHN/CIFAR-100, 
we use a batch size of 32/128 to fine-tune the network with 20/200 epochs. 
The learning rate is initialized at 0.05 and multiplied by 0.14 at 40\% and 80\% of the total number of epochs.
On ILSVRC-2012, we use a batch size of 128 to fine-tune VGG-16/ResNet-18/MobileNet-V2
for 30/100/100 epochs.
For VGG-16/ResNet-18, the learning rate is started at 0.0006 and multiplied by 0.4 at 40\% and 80\% of the total number of epochs.
We use a cosine decay learning rate schedule for MobileNet-V2~\cite{sandler2018mobilenetv2} with an initial rate of 0.03.

\input{Table/ImageNet_Transfer_Pruning_Table}

\noindent\textbf{SVHN.} We first evaluate $\xi^*$ on SVHN with ResNet-164. 
\ourmethod\ outperforms SLIM~\cite{liu2017learning} by 0.1\%  in accuracy
with significant hardware resource savings: 
32.1\% more FLOPs saving
and 48.5\% more parameters saving,
which well demonstrates the effectiveness of $\xi^*$.

\noindent\textbf{CIFAR-100.}
On VGG-19, our pruned model achieves an accuracy gain of 0.35\%
with respect to G-SD~\cite{liu2020rethinking}.
Compared to LFPC~\cite{he2020learning} and LeGR~\cite{chin2020towards}, 
our pruned ResNet-56 achieves an accuracy gain of 0.87\% and 0.66\%, respectively, while having 5\% less FLOPs.
On ResNet-110, 
our method outperforms
FPGM~\cite{he2019filter} and TAS~\cite{dong2019network} 
by 1.30\% and 0.69\% in terms of accuracy 
with 4\% less FLOPs.
In comparison with 
LCCL~\cite{dong2017more}, SLIM~\cite{liu2017learning}, 
and DI~\cite{kung2019methodical},
our pruned ResNet-164 achieves an accuracy of 77.77\% with 63.2\% FLOPs reduction which advances all prior methods.

\noindent\textbf{ILSVRC-2012.} 
On VGG-16, our approach improves over the baseline by nearly 1.1\% in top-1 accuracy with 2.4$\times$ acceleration.
Our 3.3$\times$-accelerated model advances the state of the art by achieving top-1/top-5 accuracies of 71.64\%/90.60\%. 
On ResNet-18, our approach reduces 16.8\% of the FLOPs without top-1 accuracy loss.
Compared to LCCL~\cite{dong2017more}, 
our method achieves a 2.72\% top-1 accuracy gain with a higher FLOPs reduction ratio.
We demonstrate top-1 accuracy gains of 1.75\% and 1.50\% with respect to SFP~\cite{he2018soft} and DCP~\cite{zhuang2018discrimination} with over 40\% FLOPs reduction.
We finally show our performance on a much more compact network, MobileNet-V2, 
which is specifically designed for mobile deployment.
When 26.9\% of FLOPs are pruned, our approach outperforms AMC~\cite{he2018amc}, Meta~\cite{liu2019metapruning}, and LeGR~\cite{chin2020towards} with a top-1 accuracy of 71.90\%.
At a higher pruning ratio, 
our method advances DCP~\cite{zhuang2018discrimination} and Meta~\cite{liu2019metapruning}
by top-1 accuracies of 4.94\% and 0.96\%,
with 53.4\% FLOPs reduction.

%% file: Table/ImageNet_Transfer_Pruning_Table.tex
\begin{table*}[t]
    \fontsize{8.5}{10}\selectfont
    \centering
    
    \begin{tabular}{|c|c|c|c|c|c|c|c|}
    \hline
    Network & Method & \thead{ Top-1 \\ Acc. (\%)}  & \thead{ Top-1 \\ $\downarrow$ (\%)} & \thead{Top-5 \\ Acc. (\%)} & \thead{Top-5 \\ $\downarrow$ (\%)} & \thead{FLOPs (B) \\ Pruned (\%)}  & \thead{Params (M) \\ Pruned (\%)} \\
    \hline\hline
    
    
    
    \multirow{8}{3.5em}{VGG16} & L1 \cite{li2016pruning} & - & - & 89.90 $\rightarrow$ 89.10 & 0.80 & 7.74 (50.0) & -  
    \\
    
    & CP \cite{he2017channel} & - & - &
    89.90 $\rightarrow$ 89.90 & 0.00 & 7.74 (50.0) & -  
    \\
    
    & G-SD~\cite{liu2020rethinking} & 71.30 $\rightarrow$ 71.88 & -0.58 & 90.10 $\rightarrow$ 90.66 & -0.56 & 6.62 (57.2) & 133.6 (3.4) \\
    
    & {\bf \ourmethod\ 59\%-pruned} & {\bf 71.30 $\rightarrow$ 72.37} & {\bf -1.07} & {\bf 90.10 $\rightarrow$ 91.05} & {\bf -0.95} & {\bf 6.34 (59.0)} & {\bf 133.5 (3.5)}
    \\ \cdashline{2-8}

    & RNP \cite{lin2017runtime} & - & - & 89.90 $\rightarrow$ 86.67 & 3.23 & 5.16 (66.7) & 138.3 (0.0) \\
    
    & SLIM \cite{liu2017learning} & - & - & 89.90 $\rightarrow$ 88.53 & 1.37 & 5.16 (66.7) & - 
    \\
    
    & FBS \cite{gao2018dynamic} & - & - &
    89.90 $\rightarrow$ 89.86 & 0.04 & 5.16 (66.7) & 138.3 (0.0) 
    \\
    
    & {\bf \ourmethod\ 67\%-pruned} & {\bf 71.30 $\rightarrow$ 71.64} & {\bf -0.34} & {\bf 90.10 $\rightarrow$ 90.60} & {\bf -0.50} & {\bf 5.12 (66.9)} & {\bf 131.6 (4.8)} \\ 
    \hline\hline
    
    
    \multirow{9}{4em}{\thead{ResNet \\ 18}} & \textbf{\ourmethod\ 17\%-pruned} & \textbf{70.05 $\rightarrow$ 70.08} & \textbf{-0.03} & \textbf{89.40 $\rightarrow$ 89.24} & \textbf{0.16} & \textbf{1.50 (16.8)} & \textbf{11.2 (3.9)} \\\cdashline{2-8}    
    
    & SLIM \cite{liu2017learning} & 68.98 $\rightarrow$ 67.21 & 1.77 & 88.68 $\rightarrow$ 87.39 & 1.29 & 1.31 (28.0) & - \\
    
    & LCCL \cite{dong2017more} & 69.98 $\rightarrow$ 66.33 & 3.65 & 89.24 $\rightarrow$ 86.94 & 2.30 & 1.18 (34.6) & 11.7 (0.0)
    \\
     
     
     & \textbf{\ourmethod\ 37\%-pruned} & \textbf{70.05 $\rightarrow$ 69.09} & \textbf{0.96} & \textbf{89.40 $\rightarrow$ 88.59} & \textbf{0.81} & \textbf{1.14 (36.7)} & \textbf{9.3 (20.1)}
    \\\cdashline{2-8}         
    
    & SFP \cite{he2018soft} & 70.28 $\rightarrow$ 67.10  & 3.18 & 89.63 $\rightarrow$ 87.78 & 1.85 & 1.06 (41.8) & - \\

    & DCP \cite{zhuang2018discrimination} & 69.64 $\rightarrow$ 67.35 & 2.29 & 88.98 $\rightarrow$ 87.60 & 1.38 & 0.98 (46.0) & 6.2 (47.0) \\ 
    
    & FPGM \cite{he2019filter} & 70.28 $\rightarrow$ 68.41 & 1.87 & 89.63 $\rightarrow$ 88.48 & 1.15 & 1.06 (41.8) & - \\
    
    & DSA \cite{ning2020dsa} & 69.72 $\rightarrow$ 68.61 & 1.11 & 89.07 $\rightarrow$ 88.35 & 0.72 & 1.09 (40.0) & - \\
    
     
     & {\bf \ourmethod\ 41\%-pruned} & \textbf{70.05 $\rightarrow$ 68.85} & \textbf{1.20} & \textbf{89.40 $\rightarrow$ 88.45} & \textbf{0.95} & \textbf{1.07 (41.0)} & \textbf{8.8 (24.5)}
    \\
    
    \hline\hline
        
 \multirow{9}{4em}{\thead{MobileNet \\ V2}} &  Uniform~\cite{sandler2018mobilenetv2} & 71.80 $\rightarrow$ 69.80 & 2.00 & - & - & 0.22 (26.9) & - \\

& AMC~\cite{he2018amc} & 71.80 $\rightarrow$ 70.80 & 1.00 & - & - & 0.22 (26.9) & - \\

& CC~\cite{li2021towards} & 71.88 $\rightarrow$ 70.91 & 0.89 & - & - & 0.22 (28.3) & - \\

& Meta~\cite{liu2019metapruning} & 72.70 $\rightarrow$ 71.20 & 1.50 & - & - & 0.22 (27.9) & - \\

& LeGR~\cite{chin2020towards} & 71.80 $\rightarrow$ 71.40 & 0.40 & - & - & 0.22 (26.9) & - \\

& \textbf{\ourmethod\ 27\%-pruned} & \textbf{72.18 $\rightarrow$ 71.90} & \textbf{0.28} & \textbf{90.49 $\rightarrow$ 90.38} & \textbf{0.11} & \textbf{0.22 (26.9)} & \textbf{2.8 (20.4)} \\
\cdashline{2-8}         

& DCP \cite{zhuang2018discrimination} & 70.11 $\rightarrow$ 64.22 & 5.89 & - & 3.77 & 0.17 (44.7) & 2.6 (25.9) \\

& Meta~\cite{liu2019metapruning} & 72.70 $\rightarrow$ 68.20 & 4.50 & - & - & 0.14 (53.4) & - \\

& \textbf{\ourmethod\ 53\%-pruned} & \textbf{72.18 $\rightarrow$ 69.16} & \textbf{3.02} &  \textbf{90.49 $\rightarrow$ 88.66} & \textbf{1.83} & \textbf{0.14 (53.4)} & \textbf{2.1 (39.3)} \\




\hline
        
    \end{tabular}
 \caption{ILSVRC-2012 Transfer Pruning Results. 
 We report our pruned models at different FLOPs levels to ensure a fair comparison with different prior arts.
 We add a suffix specifying FLOPs pruning percentage for each of our pruned model.
 }\label{tab:ILSVRC-2012}

\end{table*}

%% file: ablation_study.tex
\section{Ablation Study}\label{sec:ablation_study}

\begin{figure}[t]
    \centering
    \includegraphics[width=0.35\textwidth]{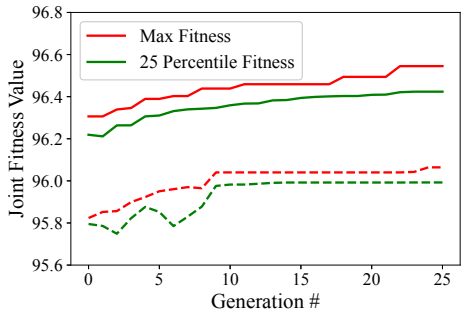}
    \vspace{-0.4cm}
    \caption{Comparing random initial population evolution (dashed line) with the evolution in Sec.~\ref{sec:co-evolve} (solid line). 
    Thanks to the expressiveness of our function space, the evolution with randomly-initialized functions also achieve good pruning fitness.
    However, we observe that it  converges very early around the 8th generation and stalls at the plateau for a long period. 
    Moreover, its final fitness has a clear performance gap with respect to the one in Sec.~\ref{sec:co-evolve}. 
    }
    \label{fig:no_SOAP_evolution}
\end{figure}

\textbf{Random Initial Population.} 
In Fig.~\ref{fig:no_SOAP_evolution}, 
we conduct a control experiment which initializes all individuals as random expression trees 
to study the effectiveness of initializing our population with SOAP.
We also turn off the SOAP function insertion in the reproduction process for the control experiment.
All other parameters (number of generations, size of population, $\alpha$, etc.) are kept to be the same as in Sec.~\ref{sec:co-evolve} for a fair comparison.
We find that evolving with random population also achieves a good pruning fitness, 
which indicates that our design space is of powerful expressiveness. 
However, we observe early convergence and a final performance gap in the control experiment compared to the main experiment in Sec.~\ref{sec:co-evolve}, 
demonstrating the advantage of using SOAP for evolution.

\noindent\textbf{Evolution on ILSVRC-2012.}
In contrast to our evolution strategy with a joint fitness function on MNIST and CIFAR-10, 
we conduct an evolution on only ILSVRC-2012 as a control experiment. 
We restrict the total computation budget to be the same as Sec.~\ref{sec:co-evolve}, i.e. 98 GPU-days, 
and evolve on ResNet-18 with a population size of 40 over 25 generations. 
Due to the constrained budget, 
each pruned net is only retrained for 4 epochs.
We include detailed evolution settings and results in Supplementary.
Two major drawbacks are found with this evolution strategy:
(1) \textbf{Imprecise evaluation.} 
Due to the lack of training epochs, the function's actual effectiveness is not precisely revealed.
We take two functions with fitness 63.24 and 63.46 from the last generation, 
and use them again to prune ResNet-18 but fully retrain for 100 epochs.
We find that the one with lower fitness in evolution achieves an accuracy of 68.27\% in the full training,
while the higher one only has an accuracy of 68.02\%. 
Such result indicates that the evaluation in this evolution procedure could be inaccurate, 
while our strategy ensures a full retraining for precise effectiveness assessment.
(2) \textbf{Inferior performance.} 
The best evolved function with this method, $\xi_{ImageNet}$ (in Supplementary), 
performs inferior to $\xi^*$ shown in Eqn.~\ref{eqn:best-func} 
when transferred to a different dataset.
In particular, when applied to pruning 56\% FLOPs from ResNet-110 on CIFAR-100,
$\xi_{ImageNet}$ only achieves an accuracy of 72.51\% while $\xi^*$ reaches 73.85\%.
These two issues suggest that evolving on two small datasets would have better cost-effectiveness than using a single large scale dataset like ILSVRC-2012.

\begin{figure}[t]
    \centering
    \includegraphics[width=0.4\textwidth]{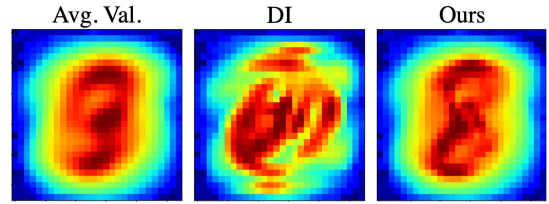}
    \vspace{-0.3cm}
    \caption{Feature selection by DI~\cite{kung2019methodical} (\textbf{middle}) and $\xi^*$ (\textbf{right}) for MNIST, 
    where $\xi^*$ tends to preserve features with higher means and more robust pattern in reference of the average feature values map (\textbf{left}). 
    }\label{fig:visual_select}
\end{figure}

\noindent\textbf{Feature Selection.} 
We further apply $\xi^*$ to another machine learning task, 
feature selection, 
to visually understand our evolved function.
In particular, we compare $\xi^*$ (\textbf{right}) vs. DI~\cite{kung2019methodical} (\textbf{middle}) on MNIST feature selection in Fig.~\ref{fig:visual_select}.
The red pixels indicate the important features evaluated by the metrics, 
while the blue ones are redundant.
Taking the average feature values map (\textbf{left}) for reference, 
we find that our evolved function tends to select features with higher means, where the MNIST pattern is more robust.

%% file: conclusion.tex
\section{Conclusion} \label{sec:conclusion}

In this work, we propose a novel paradigm integrating evolutionary learning with channel pruning, 
which first learns novel channel pruning functions from small datasets,
and then transfers them to larger and more challenging datasets. 
We develop an end-to-end genetic programming framework to 
automatically search for transferable pruning functions 
over our novel function design space without any manual modification after evolution. 
We present and analyze a closed-form evolved function which 
can offer strong pruning performance and further streamline the design of our pruning strategy.
The learned pruning function exhibits 
remarkable generalizability to datasets different from those in the evolution process.
Specifically, on SVHN, CIFAR-100, and ILSVRC-2012, we achieve state-of-the-art pruning results.


%% file: acknowledgement.tex
\section{Acknowledgement}

We thank a former colleague from Princeton Parallel Group, Yanqi Zhou, 
for her help with parallel implementation of the evolution.
This material is based upon work supported by the National Science Foundation under Grant No. CCF-1822949.  Any opinions, findings, and conclusions or recommendations expressed in this material are those of the authors and do not necessarily reflect the views of the National Science Foundation.

%% file: supplementary.tex
\newcommand\feamap{\mathcal{F}}

\input{Supplementary_Material/introduction}
\input{Supplementary_Material/SOAP_Func}

\input{Supplementary_Material/evaluation}

\input{Supplementary_Material/ImageNet_evolution}
\input{Supplementary_Material/more_evolved_func}

%% file: Supplementary_Material/introduction.tex
We organize our supplementary material as follows. 
In Sec.~\ref{sec:SOAP}, we present a more detailed table for the operator space and our implementation of the state-of-the-art population (SOAP). 
In Sec.~\ref{sec:eval_details}, we include more experimental details of our evolution and pruning study.  
We discuss more detailed settings and results of evolution on ILSVRC-2012 in Sec.~\ref{sec:imagenet_evolution}.
Lastly, we present extra evolved functions in Sec.~\ref{sec:more_evolved}.

%% file: Supplementary_Material/SOAP_Func.tex
\section{SOAP Implementation}\label{sec:SOAP}

\subsection{Operator Space}
In Tab.~\ref{tab:detailed_operator}, we present the detailed operator space with operators and their abbreviations.

\begin{table}
\centering

    \begin{tabular}{|p{2cm}|c|c|}

    \hline
    \multirow{8}{1em}{\bf Elementwise \\ operators} & addition & $\mathrm{add}(+)$ \\ \cline{2-3}
    & subtraction & $\mathrm{sub}(-)$ \\	\cline{2-3}
    & multiplication & $\mathrm{mul}(\times)$ \\	\cline{2-3}
    & division & $\mathrm{div}(\div)$ \\	\cline{2-3}
    & absolute value & $\mathrm{abs} $ \\	\cline{2-3}
    & square & $\mathrm{sq}$ \\	\cline{2-3}
    & square root & $\mathrm{sqrt}$ \\	\cline{2-3}
    & adding ridge factor & $\mathrm{ridge}$ \\	
    \hline
    
     \multirow{6}{7.5em}{\bf Matrix \\ operators} & matrix trace & $\mathrm{tr}$ \\ \cline{2-3}
     & matrix multiplication & $\mathrm{matmul}$ \\ \cline{2-3}
     & matrix inversion & $\mathrm{inv}$ \\ \cline{2-3}
     & inner product & $\mathrm{dot}$ \\ \cline{2-3}
     & outer product & $\mathrm{outprod}$ \\ \cline{2-3}
     & matrix/vector transpose  & $\mathrm{tran}$ \\ 
    \hline
    
    \multirow{6}{8.5em}{\bf Statistics \\ operators} & summation & $\mathrm{sum}_{\{s,g\}}$ \\ \cline{2-3}
    & product & $\mathrm{prod}_{\{s,g\}}$ \\ \cline{2-3}
    & mean & $\mathrm{mean}_{\{s,g\}}$ \\ \cline{2-3}
    & standard deviation & $\mathrm{std}_{\{s,g\}}$ \\ \cline{2-3}
    & variance  & $\mathrm{var}_{\{s,g\}}$ \\ \cline{2-3}
    & counting measure & $\mathrm{count}_{\{s,g\}}$  \\ 
    \hline
    
     \multirow{3}{9.5em}{\bf Specialized \\ operators} &  rbf kernel matrix getter & $\mathrm{rbf}$ \\ \cline{2-3}
     & geometric median getter & $\mathrm{geo}$ \\ \cline{2-3}
      & tensor slicer & $\mathrm{slice}$ \\ 
    \hline
    
    \end{tabular}
    \caption{Detailed Operator Space}~\label{tab:detailed_operator}
\end{table}

\subsection{SOAP Functions}

With the abbreviations of operators in Tab.~\ref{tab:detailed_operator} and the symbols of operands presented in Tab.~1 of the main paper,
we can thus give the precise expressions of the functions in SOAP:
\begin{itemize}

\item Filter's $\ell1$-norm: $\mathrm{sum}_g(\mathrm{abs}(\mathcal{W}_I))$

\item Filter's $\ell2$-norm: $\mathrm{sqrt}(\mathrm{sum}_g(\mathrm{sq}(\mathcal{W}_I)))$

\item Batch normalization's scaling factor: $\mathrm{abs}(\mathrm{slice}(\mathcal{B}))$

\item Filter's geometric median: $\mathrm{sqrt}(\mathrm{sum}_g(\mathrm{sq}(\mathcal{W}_I - \mathrm{geo}(\mathcal{W}))))$

\item Discriminant Information: 

$\mathrm{count}_s(\mathcal{F}^+)
\times
\mathrm{matmul}(
\mathrm{tran}(\mathrm{mean}_s(\mathcal{F}^+) - \mathrm{mean}_s(\mathcal{F})), \\
\mathrm{inv}(\mathrm{ridge}(\mathrm{matmul}(
\mathrm{tran}(\mathcal{F} - \mathrm{mean}_s(\mathcal{F})), 
\mathcal{F} - \mathrm{mean}_s(\mathcal{F})
))),
\\
\mathrm{mean}_s(\mathcal{F}^+) - \mathrm{mean}_s(\mathcal{F})
)
$

\item Maximum Mean Discrepancy: 

$\mathrm{div}(\mathrm{sum}_g(\mathrm{rbf}(\mathcal{F}^+, \mathcal{F}^+) ),~ \mathrm{sq}(\mathrm{count}_s(\mathcal{F}^+)))
\\
+ \mathrm{div}(\mathrm{sum}_g(\mathrm{rbf}(\mathcal{F}^-, \mathcal{F}^-)) ,~ \mathrm{sq}(\mathrm{count}_s(\mathcal{F}^-))) 
\\
- \mathrm{div}(\mathrm{sum}_g(\mathrm{rbf}(\mathcal{F}^+, \mathcal{F}^-)) ,~ \mathrm{mul}(\mathrm{count}_s(\mathcal{F}^+)), \mathrm{count}_s(\mathcal{F}^-))) -\\ \mathrm{div}(\mathrm{sum}_g(\mathrm{rbf}(\mathcal{F}^+, \mathcal{F}^-)) ,~ \mathrm{mul}(\mathrm{count}_s(\mathcal{F}^+)), \mathrm{count}_s(\mathcal{F}^-)))$

\item Generalized Absolute SNR:

$\mathrm{div}(\mathrm{abs}(\mathrm{mean}_g(\mathcal{F}^+) - \mathrm{mean}_g(\mathcal{F}^-)) ,~
\mathrm{std}_g(\mathcal{F}^+) + \mathrm{std}_g(\mathcal{F}^-)
)$

\item Generalized Student's T-Test:

$\mathrm{div}(
\mathrm{abs}(
\mathrm{mean}_g(\mathcal{F}^+) - \mathrm{mean}_g(\mathcal{F}^-)), \\
\mathrm{sqrt}(
\mathrm{div}(\mathrm{var}_g(\mathcal{F}^+), \mathrm{count}_s(\mathcal{F}^+))~+ \\
\mathrm{div}(\mathrm{var}_g(\mathcal{F}^-), \mathrm{count}_s(\mathcal{F}^-))
)
)
$

\item Generalized Fisher Discriminat Ratio:

$\mathrm{div}(\mathrm{sq}(\mathrm{mean}_g(\mathcal{F}^+) - \mathrm{mean}_g(\mathcal{F}^-)) ,~
\mathrm{var}_g(\mathcal{F}^+) + \mathrm{var}_g(\mathcal{F}^-)
)$

\item Generalized Symmetric Divergence: 

$\mathrm{div}(\mathrm{var}_g(\mathcal{F}^+), \mathrm{var}_g(\mathcal{F}^-)) + 
\mathrm{div}(\mathrm{var}_g(\mathcal{F}^-), \mathrm{var}_g(\mathcal{F}^+)) 
\\
+ \mathrm{div}(\mathrm{sq}(\mathrm{mean}_g(\mathcal{F}^+) - \mathrm{mean}_g(\mathcal{F}^-)) ,~
\mathrm{var}_g(\mathcal{F}^+) + \mathrm{var}_g(\mathcal{F}^-)
)
$

\end{itemize}

%% file: Supplementary_Material/evaluation.tex
\section{Experimental Details}\label{sec:eval_details}

\subsection{Study on Fitness Combination Scheme}

\textbf{Preliminary Evolution.} We conduct 10 preliminary experiments, where the variables are: $\alpha \in \{0, 0.3, 0.5, 0.7, 1\}$ and combination scheme $\in$ \{weighted geometric mean, weighted arithmetic mean\}.
For each experiment, we have a population of 15 functions which are evolved for 10 generations.
The population is initialized with 10 individuals randomly cloned from SOAP and 5 random expression trees.
The tournament size is 3, and the number of the selected functions is 5. 
The next generation is reproduced only from the selected functions.
Other settings are the same as the main evolution experiment.

\noindent\textbf{CIFAR-100 Pruning.} 
We apply the best evolved functions from each preliminary evolution test to prune a ResNet-38~\cite{he2016deep} on CIFAR-100~\cite{krizhevsky2009learning}.
The baseline ResNet-38 adopts the bottleneck block structure with an accuracy of 72.3\%.
We use each evolved function to prune 40\% of channels in all layers uniformly,
resulting in a 54.7\%/52.4\% FLOPs/parameter reduction.
The network is then fine-tuned by the SGD optimizer with 200 epochs.
We use the Nesterov Momentum~\cite{nesterov1983method} with a momentum of 0.9.
The mini-batch size is set to be 128, and the weight decay is set to be 1e-3.
The training data is transformed with a standard data augmentation scheme~\cite{he2016deep}.
The learning rate is initialized at 0.1 and divided by 10 at epoch 80 and 160.

\subsection{Main Evolution Experiment}
\textbf{MNIST Pruning.} On MNIST~\cite{lecun1998gradient} pruning task, 
we prune a LeNet-5~\cite{lecun1998gradient} with a baseline accuracy of 99.26\%
from shape of 20-50-800-500 to 5-12-160-40.
Such pruning process reduces 92.4\% of FLOPs and 98.0\% of parameters.
The pruned network is fine-tuned for 300 epochs with a batch size of 200 and a weight decay of 7e-5.
We use the Adam optimizer~\cite{kingma2014adam} with a constant learning rate of 5e-4.

\noindent\textbf{CIFAR-10 Pruning.} For CIFAR-10~\cite{krizhevsky2009learning} pruning, 
we adopt the VGG-16 structure from~\cite{li2016pruning} with a baseline accuracy of 93.7\%.
We uniformly prune 40\% of the channels from all layers 
resulting in 63.0\% FLOPs reduction and 63.7\% parameters reduction.
The fine-tuning process takes 200 epochs with a batch size of 128.
We set the weight decay to be 1e-3 and the dropout ratio to be 0.3.
We use the SGD optimizer with Nesterov momentum~\cite{nesterov1983method}, where the momentum is set to be 0.9.
We augment the training samples with a standard data augmentation scheme~\cite{he2016deep}.
The initial learning rate is set to be 0.006 and multiplied by 0.28 at 40\% and 80\% of the total number of epochs.

\subsection{Transfer Pruning}

We implement the pruning experiments in TensorFlow~\cite{abadi2016tensorflow} 
and carry them out with NVIDIA Tesla P100 GPUs.
CIFAR-100 contains 50,000/10,000 training/test samples in 100 classes.
SVHN is a 10-class dataset 
where we use 604,388 training images for network training with a test set of 26,032 images.
ILSVRC-2012 contains 1.28 million training images and 50 thousand validation images in 1000 classes.
We adopt the standard data augmentation scheme~\cite{he2016deep} for CIFAR-100 and ILSVRC-2012.

\subsection{Channel Scoring}

As many of our pruning functions require activation maps of the channels to determine channels' importance, 
we need to feed-forward the input images for channel scoring.
Specifically, for pruning experiments on MNIST, CIFAR-10, and CIFAR-100, 
we use all their training images to compute the channel scores.
On SVHN and ILSVRC-2012, 
we randomly sample 20 thousand and 10 thousand training images for channel scoring, respectively.

%% file: Supplementary_Material/ImageNet_evolution.tex
\section{Evolution on ILSVRC-2012}\label{sec:imagenet_evolution}

\textbf {Evolution.} We use ResNet-18 as the target network for pruning function evolution on ILSVRC-2012. 
Since only one task is evaluated, we directly use the retrained accuracy of the pruned network as the function's fitness.
Other evolution settings for population, selection, mutation, and crossover are kept the same as Sec.~4 of the main paper.

\noindent\textbf{Evaluation.} We uniformly prune 30\% of channels in each layer from a pretrained ResNet-18, resulting in a FLOPs reduction of 36.4\%.
Due to the constrained computational budget, 
we only fine-tune it for 4 epochs using the SGD optimizer with Nesterov momentum~\cite{nesterov1983method}.
We use a batch size of 128 and initialize our learning rate at 0.001.
The learning rate is multiplied by 0.4 at epoch 1 and 2.

\begin{figure}[t]
\vspace{-0.3cm}
    \centering
    \includegraphics[width=0.47\textwidth]{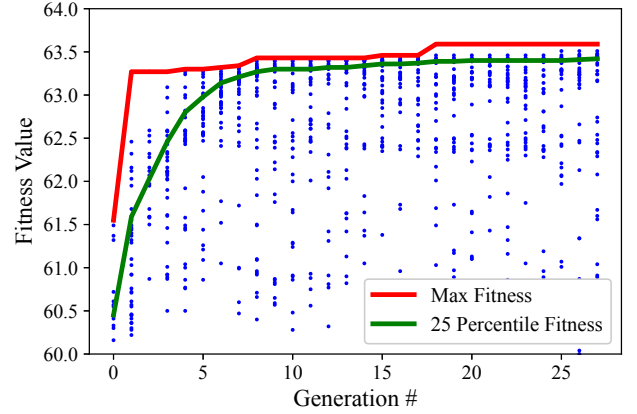}
    \caption{Function evolution on ImageNet.}
    \label{fig:evo_on_imagenet}
\end{figure}

\noindent\textbf{Result.} 
We show the evolution progress in Fig.~\ref{fig:evo_on_imagenet}.
Due to the lack of training budget, the pruned net is clearly not well retrained as they only achieve around 63.5\% accuracy, 
much lower than the performance of methods shown in Tab.~5 of the main paper at the similar pruning level.
Such inadequate training results in a imprecise function fitness evaluation evidenced in Sec.~6 of the main paper.
Moreover, the best evolved function from this strategy, $\xi_{ImageNet}$ (Eqn.~\ref{eqn:imagenet_evolved}), 
performs inferior to the co-evolved function $\xi^*$ when transferred for CIFAR-100 pruning.
These results demonstrate the advantage of our small dataset co-evolution strategy in cost-effectiveness.

\begin{table}[t]
\begin{equation}\label{eqn:imagenet_evolved}
\xi_{ImageNet}(\mathcal{C}) = (\frac{\mathrm{var}_g(\mathrm{mean}_s(\mathcal{F}^+))}{\mathrm{std}_g(\mathrm{tr}(\mathcal{F}^+)) \times \mathrm{mean}_g(\mathcal{F}^-)})^4 \div \mathrm{var}_g(\mathrm{sqrt}(\mathcal{F}))
\end{equation}
\end{table}

%% file: Supplementary_Material/more_evolved_func.tex
\section{Extra Evolved Functions}\label{sec:more_evolved}

We present additional evolved functions from our co-evolution strategy:
\begin{equation}\label{eqn:evolved_1}
    \xi_1(\mathcal{C}) = \frac{||\bar{f} - \mathrm{var}_g(\mathcal{F}^-)\mathbf{1}||_2^2}
    {\mathrm{var}_g(\mathcal{F}^+) + \mathrm{var}_g(\mathcal{F}^-)} 
    + \mathrm{var}_g(\mathcal{F}^+)
\end{equation}
\begin{equation}\label{eqn:evolved_2}
\xi_2(\mathcal{C}) = \mathrm{var}_g(\mathcal{F}^+)
\end{equation}
\begin{equation}\label{eqn:evolved_3}
\xi_3(\mathcal{C}) = \mathrm{var}_g(\mathcal{W}_I)
\end{equation}

Eqn.~\ref{eqn:evolved_1} presents a metric with the concept of SNR for classification, while having a novel way of statistics combination.
Moreover, our evolution experiments find that measuring the variance across all elements in $\mathcal{F}^+$ (Eqn.~\ref{eqn:evolved_2}) and $\mathcal{W}_I$ (Eqn.~\ref{eqn:evolved_3}) would help us identify important channels empirically.
These two functions are simple and effective yet remain undiscovered from the literature.




